\documentclass{article} 
\usepackage[preprint]{colm2025_conference}

\usepackage{microtype}
\usepackage{hyperref}
\usepackage{url}
\usepackage{booktabs}
\usepackage{caption}

\usepackage{lineno}

\usepackage{multirow}
\usepackage{adjustbox}
\usepackage{wrapfig}  
\usepackage{fontawesome5}

\definecolor{darkblue}{rgb}{0, 0, 0.5}
\hypersetup{colorlinks=true, citecolor=darkblue, linkcolor=darkblue, urlcolor=darkblue}

\title{Personalized Scientific Figure Caption Generation: An Empirical Study on Author-Specific Writing Style Transfer}

\author{Jaeyoung Kim\thanks{ Corresponding author.} , Jongho Lee \& Hongjun Choi \\
Teamreboott Inc.\\
Seoul, Korea \\
\texttt{\{jaeyoungkim,jongho.lee,hongjun.choi\}@reboott.ai} \\
\And
Sion Jang \\
MIRI D.I.H Inc. \\
Seoul, Korea \\
\texttt{wayterren@gmail.com} \\
}

\begin{document}


\maketitle

\begin{abstract}
We study personalized figure caption generation using author profile data from scientific papers. Our experiments demonstrate that rich author profile data, combined with relevant metadata, can improve the personalization performance of multimodal large language models. However, we also reveal a fundamental trade-off between matching author style and maintaining caption quality. Our findings offer valuable insights and future directions for developing practical caption automation systems that balance both objectives. This work was conducted as part of the 3rd SciCap challenge.~\footnote{\url{https://scicap.ai}}
\end{abstract}

\begin{center}
\faIcon{github}~\href{https://github.com/teamreboott/rb-ai-3rd-scicap}{\texttt{https://github.com/teamreboott/rb-ai-3rd-scicap}}
\end{center}

\begin{figure}[h]
    \centering
    \includegraphics[width=11cm]{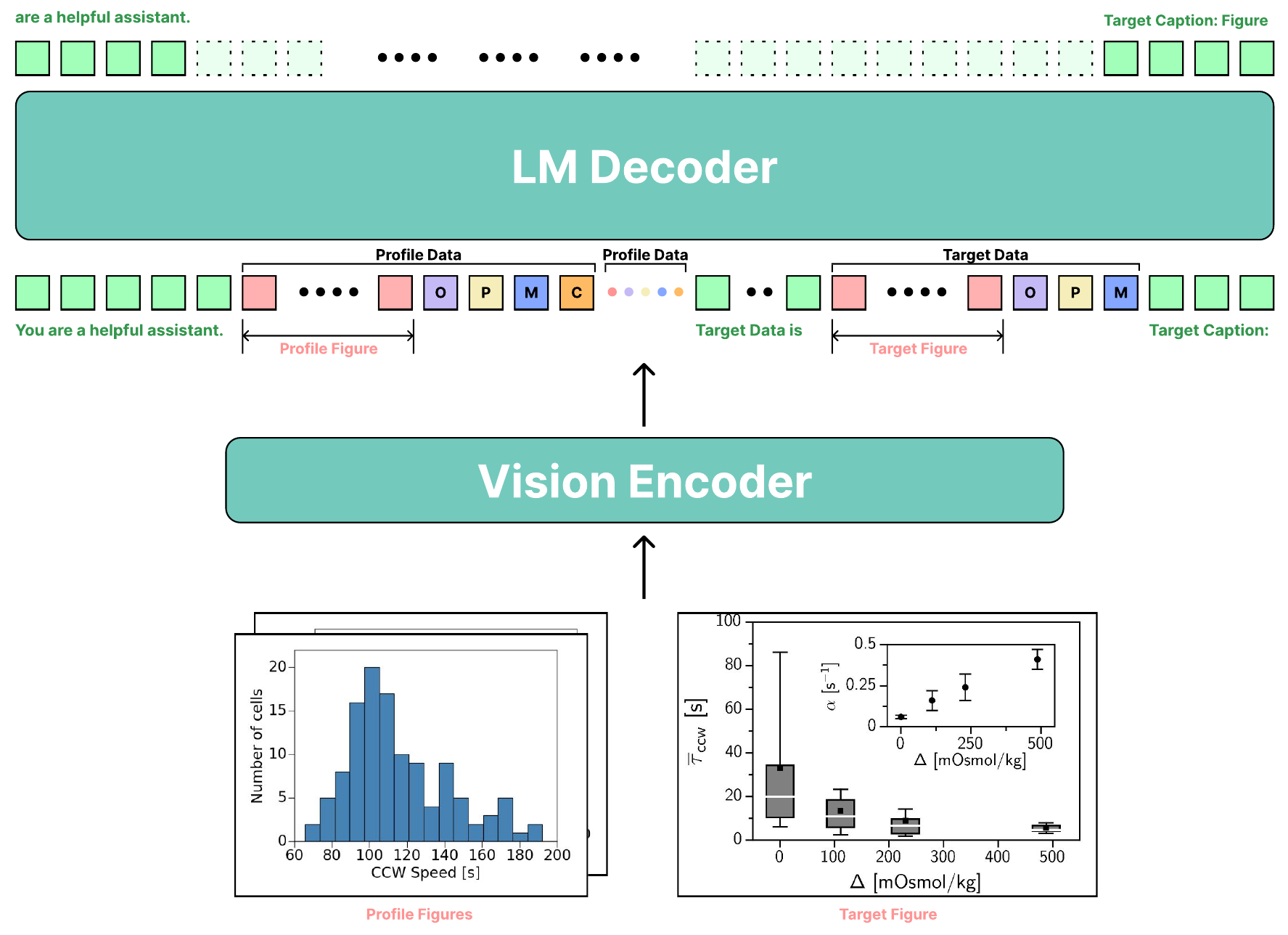}
    \caption{Multimodal LLM architecture for the personalized figure caption generation task. The model receives two types of inputs: (1) target figure information including the figure itself (F), explanatory paragraphs (P), textual mentions (M), and OCR-extracted text (O), and (2) optional profile data from N related figures in the same paper (where N can be 0), each containing corresponding figures, paragraphs, mentions, OCR text, and captions (C).}
    \label{fig:vlm_model}
\end{figure}

\section{Personalized Caption Generator} 
We propose a simple two-stage approach for personalized figure caption generation: first, we develop a quality evaluator to filter low-quality captions from training data, then we fine-tune a multimodal LLM with author-specific profile data to generate personalized captions.

\noindent \textbf{Caption Quality Evaluator.}
Previous study has demonstrated that over 50\% of figure captions in arXiv papers fail to provide meaningful information to readers \citep{huang2023summaries}. Moreover, filtering low-quality captions has been shown to improve model performance during fine-tuning \citep{kim2025multillmcollaborativecaptiongeneration}.
\begin{wrapfigure}{r}{0.4\textwidth}  
    \centering
    \includegraphics[width=\linewidth]{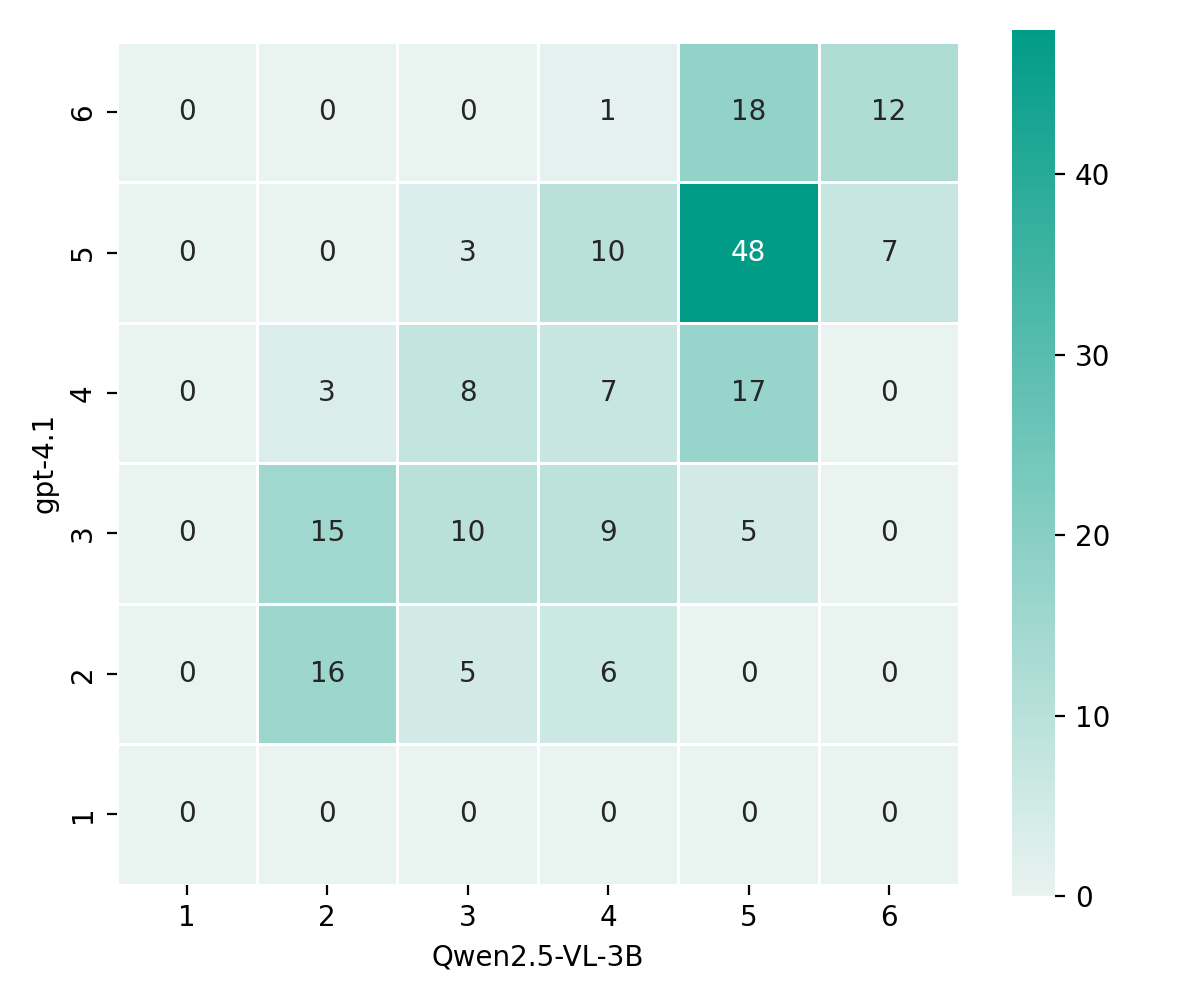}
    \caption{Confusion matrix showing the agreement between fine-tuned Qwen-2.5-VL-3B and GPT-4.1.}
    \label{fig:confusion_matrix}
\end{wrapfigure}
Following the methodology proposed by \cite{kim2025multillmcollaborativecaptiongeneration}, we develop a caption quality evaluator $f_{quality}$ to assess and filter caption quality. To train this evaluator, we first construct a synthetic quality assessment dataset using GPT-4.1~\footnote{\url{https://openai.com/index/gpt-4-1}}. Specifically, we provide GPT-4.1 with figure-caption pairs and instruct it to assign quality scores ranging from 1 (low-quality) to 6 (high-quality) points.

The validity of this scoring approach has been established in prior research, which showed that LLM-generated quality scores on SciCap-Eval~\citep{hsu2023gpt} achieved a correlation coefficient of 0.5 with doctoral student evaluations.

We curated 3,000 synthetic training samples from LaMP-Cap train dataset~\citep{ng2025lamp} and fine-tuned Qwen-2.5-VL-3B~\citep{Qwen2.5-VL} as $f_{quality}$. The fine-tuned model achieves strong agreement with GPT-4.1-generated scores on a held-out validation set of 200 samples, yielding a Spearman's rank correlation coefficient of 0.759 and a Quadratic Weighted Kappa (QWK) score of 0.754. The confusion matrix visualizing the agreement between the models is presented in Figure~\ref{fig:confusion_matrix}.

\noindent \textbf{Multimodal Large Language Model.} We adopt Qwen-2.5-VL-7B~\citep{Qwen2.5-VL} as our caption generator $g_{caption}$. To enable personalized caption generation, we provide the model with two types of inputs: the target figure along with its associated information, and profile data extracted from the same source paper (see Figure~\ref{fig:vlm_model}).
The target figure $F_{target}$ is accompanied by: (1) paragraphs $P_{target}$ containing explanations of the target figure, (2) direct textual mentions $M_{target}$ that reference the target figure, and (3) OCR-extracted text $O_{target}$ from the target figure itself. Additionally, we augment this with $N$ profile data for personalization, which includes: (1) other related figures $\{F_{profile}^i\}_{i=1}^N$ from the same paper, (2) their corresponding explanatory paragraphs $\{P_{profile}^i\}_{i=1}^N$, (3) their textual mentions $\{M_{profile}^i\}_{i=1}^N$, and (4) their OCR-extracted text $\{O_{profile}^i\}_{i=1}^N$.

For supervised fine-tuning, we used LaMP-Cap train dataset with quality scores of 3 or higher, as predicted by $f_{quality}$. The $g_{caption}$ is fine-tuned for one epoch using the AdamW optimizer~\citep{loshchilov2017decoupled} with a learning rate of $2e-5$ and a batch size of 2.

\section{Experimental Results}

\begin{table}[h]
\centering
\renewcommand{\arraystretch}{1.3}
\adjustbox{width=10cm}{
\begin{tabular}{@{}lccccccccc@{}}
\toprule
 & & \multicolumn{4}{c}{\textbf{BLEU}} & \multicolumn{3}{c}{\textbf{Rouge}} \\
\cmidrule(lr){3-6} \cmidrule(lr){7-9}
\textbf{LLM} & \textbf{\# Profile} & \textbf{BLEU-1} & \textbf{BLEU-2} & \textbf{BLEU-3} & \textbf{BLEU-4} & \textbf{Rouge-1} & \textbf{Rouge-2} & \textbf{Rouge-L} \\
\midrule
\multirow{4}{*}{Gemini-2.5-flash} 
 & 1 (n=5950) & 0.423 & 0.346 & 0.287 & 0.250 & 0.574 & 0.388 & 0.515   \\
 & 2 (n=2896) & 0.457 & 0.382 & 0.323 & 0.284 & 0.607 & 0.429 & 0.550   \\
 & 3 (n=3424) & 0.494 & 0.420 & 0.362 & 0.322 & 0.632 & 0.460 & 0.576   \\
 & All (n=12270) & 0.451 & 0.375 & 0.316 & 0.278 & 0.598 & 0.418 & 0.540   \\ \hline
 \multirow{4}{*}{Qwen2.5-VL-7B} 
 & 1 (n=5950) & 0.441 & 0.372 & 0.317 & 0.281 & 0.589 & 0.416 & 0.539   \\
 & 2 (n=2896) & 0.467 & 0.400 & 0.346 & 0.310 & 0.611 & 0.446 & 0.561   \\
 & 3 (n=3424) & 0.494 & 0.428 & 0.376 & 0.339 & 0.629 & 0.469 & 0.580   \\
 & All (n=12270) & 0.462 & 0.394 & 0.340 & 0.304 & 0.605 & 0.438 & 0.555   \\
\bottomrule
\end{tabular}
}
\caption{Performance comparison of different LLMs across multiple evaluation metrics.}
\label{tab:performance_comparison}
\end{table}

\begin{table}[h]
\centering
\renewcommand{\arraystretch}{1.5}
\adjustbox{width=10cm}{
\begin{tabular}{@{}llccccccc@{}}
\toprule
 & & \multicolumn{4}{c}{\textbf{BLEU}} & \multicolumn{3}{c}{\textbf{Rouge}}  \\
\cmidrule(lr){3-6} \cmidrule(lr){7-9}
\textbf{Profile} & \textbf{Target} & \textbf{BLEU-1} & \textbf{BLEU-2} & \textbf{BLEU-3} & \textbf{BLEU-4} & \textbf{Rouge-1} & \textbf{Rouge-2} & \textbf{Rouge-L}  \\
\midrule
F + C & F & 0.286 & 0.194 & 0.150 & 0.123 & 0.406 & 0.201 & 0.355   \\
F + P + C & F + P & 0.452 & 0.384 & 0.330 & 0.295 & 0.595 & 0.427 & 0.545   \\
F + P + M + C & F + P + M & 0.459 & 0.391 & 0.337 & 0.301 & 0.602 & 0.435 & 0.553   \\
F + P + M + O + C & F + P + M + O & \textbf{0.462} & \textbf{0.394} & \textbf{0.340} & \textbf{0.304} & \textbf{0.605} & \textbf{0.438} & \textbf{0.555}  \\
\bottomrule
\end{tabular}
}
\caption{Performance evaluation of different input configurations.}
\label{tab:profile_performance}
\end{table}

We evaluate the caption generation model on the personalized Scicap test set to assess the impact of personalization through profile data integration. Our evaluation employs standard metrics for text generation: BLEU and ROUGE scores. The ROUGE scores reported in our experiments represent F1-scores. Unless otherwise specified, all experiments are evaluated on the LaMP-Cap test dataset.

\noindent \textbf{Impact of Profile Data Quantity.} Table~\ref{tab:performance_comparison} shows the relationship between the number of profiles. We compare the fine-tuned Qwen2.5-VL-7B against Gemini-2.5-flash across test samples with varying numbers of available profile figures (1, 2, or 3). Both models show consistent performance improvements as more profile data becomes available. Notably, samples with 3 profile figures achieve the highest scores across all metrics, indicating that richer contextual information enables more accurate personalized caption generation. Interestingly, the fine-tuned Qwen2.5-VL-7B outperforms Gemini-2.5-flash on the complete test set, achieving BLEU-4 of 0.304 versus 0.278 and ROUGE-L of 0.555 versus 0.540.

\noindent \textbf{Ablation Study on Input Components.} Table~\ref{tab:profile_performance} presents an ablation study examining the contribution of each input component. Starting from a baseline configuration using only figures and captions (F+C), we progressively add paragraphs (P), mentions (M), and OCR text (O). The results reveal substantial improvements when incorporating contextual information: adding paragraphs increases BLEU-4 from 0.123 to 0.295 and ROUGE-L from 0.355 to 0.545. The addition of mentions and OCR text provides incremental but meaningful gains, with the complete configuration achieving BLEU-4 of 0.304 and ROUGE-L of 0.555. These findings validate our comprehensive approach to context integration, demonstrating that each component contributes to generating more accurate and author-centric captions.

\section{Discussion}

\noindent \textbf{Trade-off Between Personalization and Caption Quality.} Our experiments demonstrate that author-specific profile data is effective for mimicking an author's writing style, as evidenced by consistent gains in BLEU and ROUGE metrics. However, these metrics inherently prioritize lexical overlap with ground-truth captions, which may inadvertently favor style imitation over informativeness. In real-world scenarios, users of automated caption generation systems require not only stylistic consistency but also high factual and descriptive quality. This observation suggests that style and quality may occupy partially orthogonal subspaces in the model’s representation space. A model optimized solely for style matching may underrepresent quality-related features, while a model optimized purely for quality may neglect style fidelity.

\begin{figure}[h]
\centering
\begin{minipage}{0.4\textwidth}
    \centering
    \includegraphics[width=\linewidth]{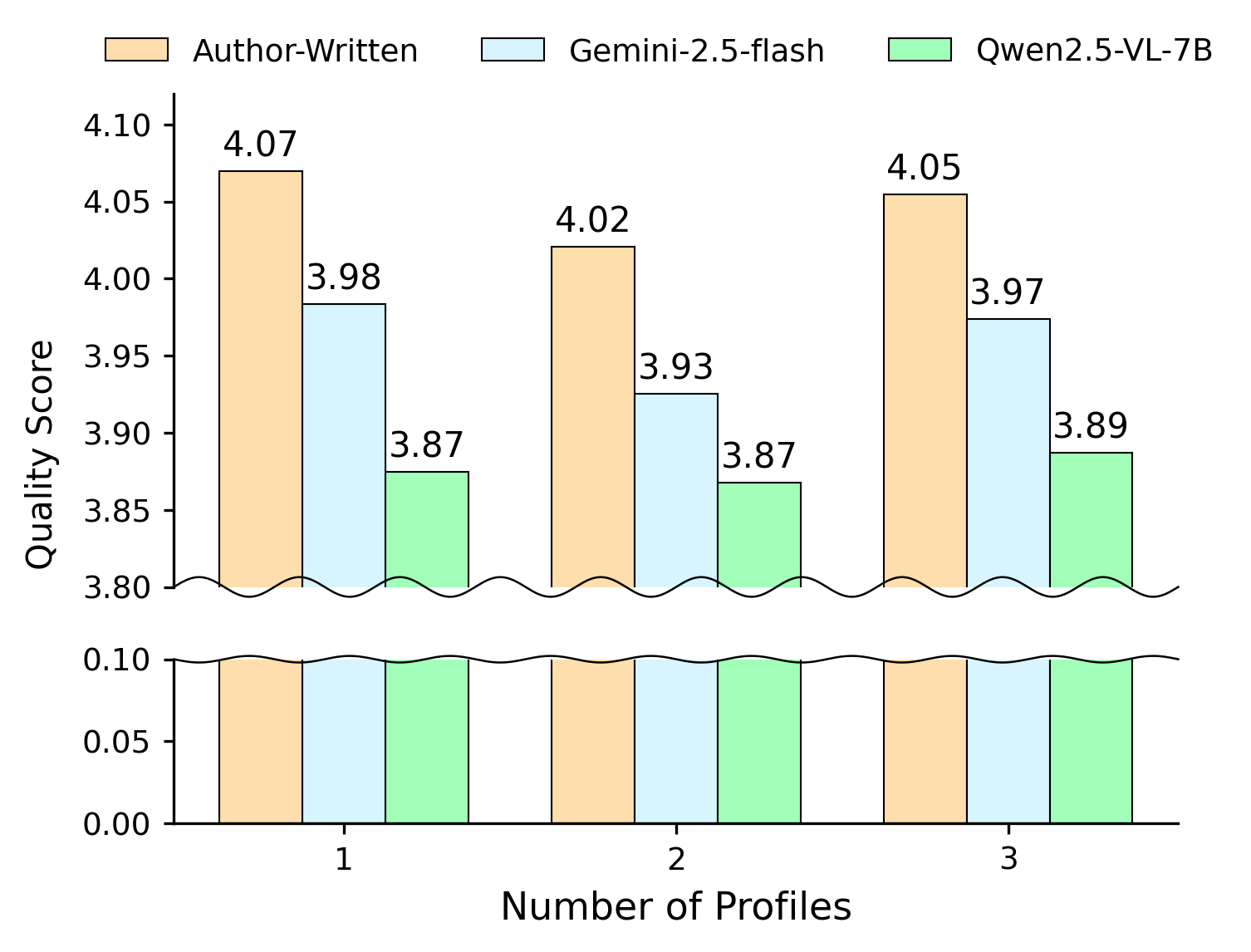}
    \caption{Caption quality score comparing Gemini-2.5-flash and fine-tuned Qwen-2.5-VL-7B. Both models generate lower quality captions compared to author-written captions.}
    \label{fig:quality_comparison}
\end{minipage}
\hfill
\begin{minipage}{0.55\textwidth}
    \centering
    \renewcommand{\arraystretch}{1.2}
    \adjustbox{width=\linewidth}{
    \begin{tabular}{@{}ccccc@{}}
    \toprule
    \multirow{2}{*}{\textbf{Target}} & \multicolumn{2}{c}{\textbf{Quality Score}} & \multicolumn{2}{c}{\textbf{ROUGE-L}} \\
    \cmidrule(lr){2-3} \cmidrule(lr){4-5}
    \textbf{Quality} & \textbf{Forced-Q6} & \textbf{Predicted-Q} & \textbf{Forced-Q6} & \textbf{Predicted-Q} \\
    \midrule
    1 & \textbf{3.743} & 3.410 & 0.469 & \textbf{0.498} \\
    2 & \textbf{3.423} & 3.236 & 0.535 & \textbf{0.571} \\
    3 & \textbf{3.582} & 3.522 & 0.571 & \textbf{0.574} \\
    4 & \textbf{4.000} & 3.962 & \textbf{0.553} & 0.548 \\
    5 & \textbf{4.435} & 4.406 & \textbf{0.530} & 0.522 \\
    6 & \textbf{4.956} & 4.917 & \textbf{0.509} & 0.502 \\
    \bottomrule
    \end{tabular}
    }
    \captionof{table}{Comparison of quality-aware caption generation strategies. Forced-Q6 conditions the model on maximum quality score (6) during inference, while Predicted-Q uses the model's own quality predictions. Target Quality indicates the quality scores of original author-written captions as evaluated by GPT-4.1.}
    \label{tab:quality_aware}
\end{minipage}
\end{figure}

To investigate this trade-off, we analyzed the quality score of generated captions using $f_{quality}$. As shown in Figure~\ref{fig:quality_comparison}, both Gemini-2.5-flash and the fine-tuned Qwen-2.5-VL-7B produce captions with moderate quality scores averaging around 3 points. Surprisingly, despite training Qwen-2.5-VL-7B without low-quality samples (quality score $\leq$ 2), it generates slightly lower quality captions compared to Gemini-2.5-flash. This phenomenon suggests that the fine-tuning process optimizes primarily for matching target caption tokens rather than maintaining caption quality, revealing an inherent tension between personalization and quality objectives.

\noindent \textbf{Quality-Aware Caption Generation.} To address the identified trade-off, we introduce a quality-aware training paradigm inspired by multi-task learning principles, where the model is jointly trained to (1) predict the expected quality score of a target caption and (2) generate the caption itself (see Figure~\ref{fig:quality_aware_model} in the Appendix). The hypothesis is that predicting quality before generation forces the model to encode quality-related signals into its latent representations, thereby creating a controllable generation process where style and quality can be jointly optimized.
This setup enables us to control the quality–personalization balance at inference time: conditioning on a high quality score (e.g., Q=6) can steer the model toward producing more informative captions, while still leveraging profile-based style conditioning. In contrast, the baseline model without explicit quality prediction lacks a mechanism to adjust this balance post-training.

Table~\ref{tab:quality_aware} compares two inference strategies: Predicted-Q, where the model generates captions conditioned on its internally predicted quality score, and Forced-Q6, where the quality score is fixed at the maximum value (6). Forced-Q6 consistently yields higher quality scores, validating that the model has learned a controllable quality dimension. 

However, in cases where the ground-truth captions are themselves of low quality, ROUGE-L scores decrease under Forced-Q6. This may be an artifact of overlap-based metrics: if the ground-truth contains omissions or vague wording, producing a more informative caption necessarily reduces n-gram overlap, leading to lower ROUGE-L despite objectively higher quality. This finding underscores the limitation of relying solely on ROUGE/BLEU for personalized scientific text generation, and motivates the need for complementary evaluation metrics that separately measure style alignment and factual/semantic quality.

\section{Conclusion} 
In this paper, we conducted an empirical study on personalized figure caption generation using author-specific profile data from scientific papers. Experimental results on the LaMP-Cap dataset demonstrate that increasing the amount of profile data consistently improves BLEU/ROUGE metrics, with the fine-tuned Qwen-2.5-VL-7B achieving competitive performance against larger models. However, our analysis reveals a trade-off between personalization and caption quality, highlighting that optimizing for author style matching alone may compromise the informativeness of generated captions. This work provides insights into personalized scientific text generation while identifying critical challenges that must be addressed to create truly practical caption automation systems that serve the dual goals of maintaining author voice and ensuring high-quality scientific communication.

\bibliography{colm2025_conference}
\bibliographystyle{colm2025_conference}

\appendix
\section{Appendix}

\subsection{Quality-Aware Caption Generation}

\begin{figure}[h]
    \centering
    \includegraphics[width=14cm]{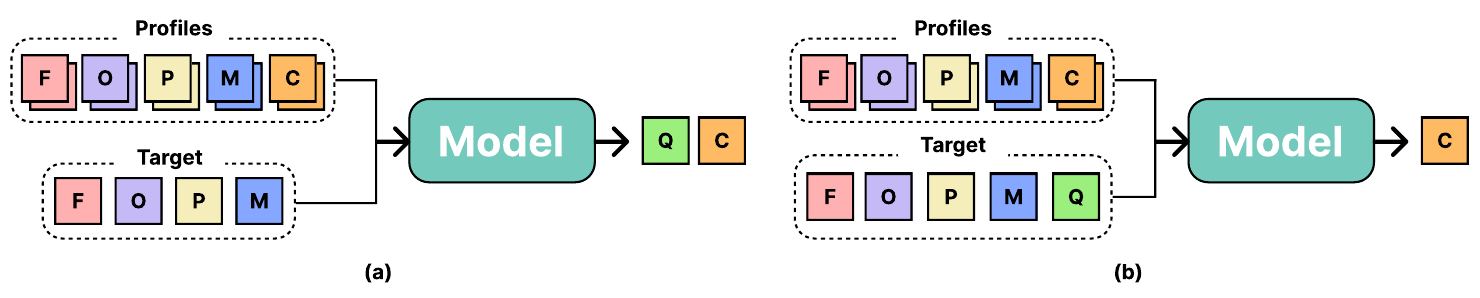}
    \caption{Quality-aware model architecture for personalized caption generation. (a) Training phase: The model predicts both the quality score (Q) and caption (C) from profile data and target figure information, learning to associate quality levels with caption characteristics. (b) Inference phase: The model can be conditioned on a specified quality score (e.g., Q=6 for maximum quality) to control the quality-personalization trade-off during generation. Profile data includes figures (F), paragraphs (P), mentions (M), OCR text (O), and captions (C) from related figures in the same paper.}
    \label{fig:quality_aware_model}
\end{figure}

To address the inherent trade-off between personalization and caption quality identified in our main experiments, we developed a quality-aware training paradigm that enables the model to jointly learn caption quality assessment and generation. This approach modifies the standard fine-tuning procedure to incorporate explicit quality signals during both training and inference.

During training, we augment the original task by requiring the model to first predict the quality score (ranging from 1 to 6) of the target caption before generating it. Specifically, given the profile data $\{F_{profile}^i, P_{profile}^i, M_{profile}^i, O_{profile}^i, C_{profile}^i\}_{i=1}^N$ and target figure information $(F_{target}, P_{target}, M_{target}, O_{target})$, the model is trained to: (1) Predict the quality score $q_{pred}$ of the ground-truth caption (2) Generate the caption $C_{pred}$ conditioned on the predicted quality.

This dual-objective training encourages the model to develop internal representations that explicitly encode quality characteristics alongside author-specific style patterns. The quality scores used for training supervision are obtained from our quality evaluator $f_{quality}$, which was validated against GPT-4.1 assessments.

The quality-aware model uses Qwen-2.5-VL-7B, fine-tuned for one epoch using AdamW with a learning rate of $2e-5$ and a batch size of 2.

At inference time, the quality-aware architecture enables two distinct generation strategies: (1) Predicted Quality (Predicted-Q): The model first predicts the likely quality score based on the input context, then generates a caption conditioned on this prediction. This approach maintains consistency with the training distribution.
(2) Forced High Quality (Forced-Q6): We override the quality prediction and force the model to condition on the maximum quality score (6), instructing it to generate the highest quality caption possible given the profile constraints.

\subsection{Prompts}

\begin{figure}[h]
    \centering
    \includegraphics[width=8cm]{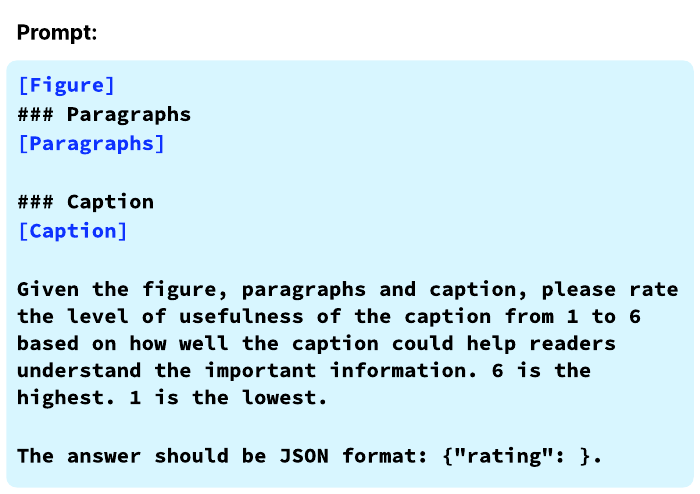}
    \caption{The actual prompt for caption quality assessment we used in our experiments.}
    \label{fig:quality_prompt}
\end{figure}

\begin{figure}[h]
    \centering
    \includegraphics[width=8cm]{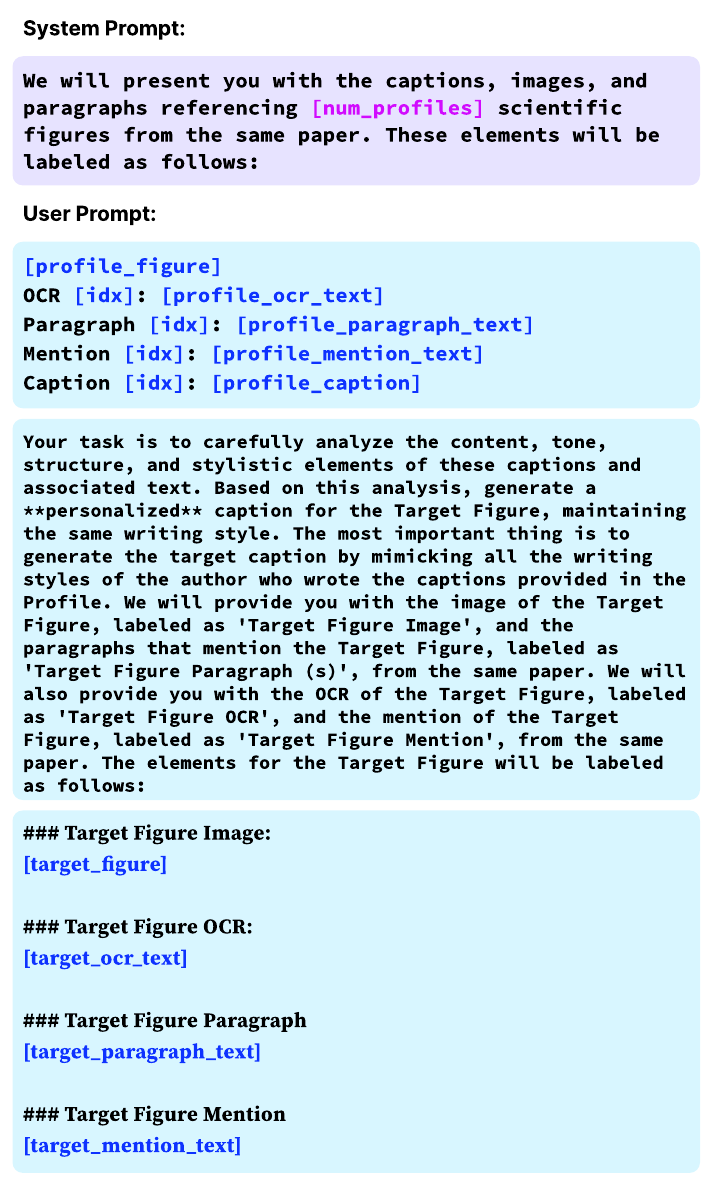}
    \caption{The actual prompt for caption generation in our experiments.}
    \label{fig:quality_prompt}
\end{figure}

\end{document}